# Are complex systems hard to evolve?

Andy Adamatzky and Larry Bull

University of the West of England, Bristol, UK

{andrew.adamatzky, larry.bull} @uwe.ac.uk

**Abstract**

Evolutionary complexity is here measured by the number of trials/evaluations needed for evolving a logical gate in a non-linear medium. Behavioural complexity of the gates evolved is characterised in terms of cellular automata behaviour. We speculate that hierarchies of behavioural and evolutionary complexities are isomorphic up to some degree, subject to substrate specificity of evolution and the spectrum of evolution parameters.

## 1. Introduction

There are many ways to measure the complexity of a system [Papadimitriou and Steiglitz, 1998; Wagner and Wechsung, 2001; Li and Vitányi, 1997]. For example, having a problem we can estimate its complexity as the number of steps a computer takes to solve the problem (computational complexity). A descriptive complexity of a system is evaluated by the length of a minimal program/model representing the system. Complexity of a finite state machine can be classified in sets of regular languages accepted by the machine. There are also measures based on how difficult it is to predict the evolution of a system.

There is however few – if any – results related to the comparison of different measures of complexity. The present short paper aims to raise readers' awareness of the subject and to invite them to undertake research in 'comparative complexity'. We compare the evolutionary complexity of logical functions (time required to evolve the functions) with the behavioural complexity of the functions (complexity of sequences of data iteratively processed by the functions).

Non-linear spatial extended media – liquid crystal [Harding and Miller, 2005] and Belousov-Zhabotinsky system [Toth et al., 2007] – were chosen as substrates of various logical gates' evolution because these media are possibly the best prototypes of non-classical, unconventional, computing devices [Adamatzky, 2001]. One-dimensional cellular automata were then selected as substrates to implement the gates evolved because the automata are good approximators of the real world [Ilachinski, 2001], the space-time dynamic of the automata is easy to visualize, and methods for exploring complexity of the automata are well-established [Wuensche and Lesser, 1992; Wolfram, 1986; Gilman, 1987; Wolfram, 1996; Martin, 1997].

## 2. Evolutionary complexity of Boolean gates

We infer hierarchies of evolutionary complexity from computational and experimental laboratory studies of evolving of logic gates in liquid crystal [Harding and Miller, 2005] and excitable chemical medium [Toth *et al.*, 2007]. See full details in the papers cited, here we provide only a very brief overview of the results.

Harding and Miller [2005] performed experiments on evolving OR, AND, NAND, NOR and XOR gates in liquid crystal. They used the main component of a liquid crystal display as the basic unit of evolving material. Eight external inputs were connected to some of 64 possible inputs of the liquid crystal. A genotype represented which of the 64 crystal inputs received external inputs (ground, two voltages representing arguments of evolved logical function, a voltage representing output of the function, four voltages to modify physical structure of the crystal, and configurations of the inputs). A fitness of any particular internal configuration of liquid crystal was evaluated by applying values of arguments of an evolved function and checking whether the response of the crystal corresponded to the value of the function. Populations of 40 individual genotypes were processed with a mutation rate of five random changes to a genotype per individual. The five fittest genotypes were selected for the next generation. The evolution ran for 200 generations. The characteristics of the evolutionary process for each of logical gates evolved are shown in Fig. 1a. These results by Harding and Miller [2005] lead us to the following ('◄ 'means 'quicker to evolve than'):

*Finding 1.* Logical gates evolved in liquid crystal have the following hierarchies of complexity: -

{OR, NOR} ◄ AND ◄ NOT ◄ NAND ◄ XOR (based on minimal number of evaluations)

NOT ◄ XOR ◄ OR ◄ NAND ◄ NOR ◄ AND (average and max. number of evaluations). □

Toth *et al.* [2007] evolved Boolean gates AND, NAND and XOR in a hybrid system "cellular automaton – excitable chemical medium". A light-sensitive modification of the Belousov-Zhabotinsky reaction was employed. A two-dimensional finite automaton controller represented a heterogeneous check-board pattern of illumination, which tuned propagation of excitation wave-fragments in the medium as they collided.

Two domains of the excitable medium were designated inputs, through which appropriate excitation was fed to the medium (e.g., low = logical '0'). A threshold of integral activity of the medium was set to differentiate between output logical values. An individual genotype in the evolving population represented the set of automata transition rules for one cell, each existing within a two-dimensional automata (10 by 10 cells) configuration. The input to each cell was the threshold activity of itself and surrounding neighbours and its output represented the illumination controlling chemical medium to be projected in that cell. The mutation rate was set either at 4000 or 6000 random alterations within the 100 automata rule tables. The fitness for each input pattern was estimated after 25 iterations. Longevity of the evolution process achieved in the chemical laboratory conditions is characterised in Fig. 1b. The table indicates that for the high mutation rate the following proposition takes place:

*Finding 2*. Logic gates evolved in light-sensitive excitable chemical medium have the following hierarchy of evolutionary complexity:

AND ◄ NAND ◄ XOR (hierarchy is the same for minimal and average evaluations)

AND ◄ {NAND, XOR} (based on maximal number of evaluations). □

## 3. Behavioural complexity of Boolean gates

We here employ one-dimensional binary cell-state cellular automata to evaluate behavioural complexity. For each particular Boolean binary operator GATE∈{AND, OR, XOR, NAND, NOR} we assign a cellular automaton where each cell $x_i$ updates its state in discrete time $t$ depending on states of its 'left' $x_{i-1}$ and 'right' $x_{i+1}$ neighbours by the rule: $x_i^{t+1}$ = GATE( $x^t_{i-1}$ , $x^t_{i+1}$ ). Examples of space-time configurations are shown in Fig.2.

The complexity of binary-neighbourhood and binary-state one-dimensional automata is very well studied [Wuensche and Lesser, 1992; Wolfram, 1996]. Behavioural and computational measures of cellular automata complexity were suggested in [Wolfram, 1986; Gilman, 1987; Cattaneo, 1999], including a more physics-related apparent entropy measure [Martin, 1997].

To keep our discussion self-consistent we will not refer to any external sources but employ a few of the simplest measures of complexity: length of transient period, morphology of configurations, and frequency of neighbourhood states.

Longer transient periods, richer (in terms of neighbourhood states occurring at every step of automaton development) morphology of space-time configurations are attributed to cellular automata at the higher levels of complexity.

Any activity in automaton with rule AND became extinct after a very short transient period (Fig. 2a). Cellular automaton governed by rule OR quickly falls into a fixed point of evolution, where every cell has a state of logical TRUTH (Fig. 2b). Thus the local transition rules AND and OR occupy lower level of behavioural complexity.

Behaviour of cellular automata controlled by NOR and NAND gates is somewhat more 'sophisticated'. Automata with NAND transitions exhibit globally periodic, configurations, with synchronized activities of their cells. The cellular lattice switches between almost everywhere TRUTH and almost everywhere FALSE global states with few domains (stationary localizations, or 'still lives' in Game of Life terminology) of cells in stationary states of FALSE bounded by states of TRUTH (Fig.2d). A cellular automaton, governed by rule NOR, just switches between two configurations: everywhere TRUTH and everywhere FALSE (Fig. 2e). The presence of at least two different global configurations in automata, ruled by NOR, places them above AND- and OR-automata in the hierarchy of behavioural complexity. NAND-automata occupy an even higher level of complexity because they exhibit a few different domains in configurations at the same time steps (Fig. 3b).

A cellular automaton, whose cell-state transition rule is XOR, has a very long and rich morphology of space-time dynamics, due to travelling and interacting defects and localizations (Fig. 2c), very long transient period (Fig. 2c), and irregular profiles of neighbourhood states frequencies (Fig. 3a). This automaton takes the highest level in the hierarchy of behavioural complexity as defined here:

*Finding 3.* Logical gates, when interpreted as local transition functions of one-dimensional binary cellular automata, have the following hierarchy of behavioural complexity:

OR ◂ NOR ◂ AND ◂ NAND ◂ XOR .

**Discussion**

Are complex systems harder to evolve? Yes. We have affirmed this by comparing results of several laboratory experiments and analysis of cellular automaton development.

Hierarchies of evolutionary complexity, derived from studies on evolving logical gates in liquid crystal [Harding and Mills, 2005] and Belousov-Zhabotinsky medium [Toth et al., 2007] are consistent in the majority of cases studied. There is a direct correlation between the evolutionary hierarchy, based on minimal number of evaluations (Findings 1 and 2.) and the behavioural complexity hierarchy (Finding 3).

The results from the liquid crystal contradict the hierarchy with respect to the average and maximal number of evaluations. However, Harding and Miller [2005] do not discuss parameter optimisation for their simulation evolutionary process. In the Belousov-Zhabotinsky case, Toth et al. [2007] show how different rates of mutation within the evolution can have effects. With a higher mutation rate, correspondence is found also found for the maximal number of evaluations with an ambiguous tie between NAND and XOR for the average due to an experimental duration cut-off point of 2000 (Finding 2.). With a lower mutation rate, the same tie is again seen but the result for the maximal duration has XOR and NAND reversed in the hierarchy.

The isomorphy between evolutionary and behavioural complexities was confirmed by experiments with evolving gates in excitable chemical medium (Finding 4). However in the case of Belousov-Zhabotinsky reaction only few of basic gates were considered as target of the evolution [Toth et al., 2007] so full comparison of evolutionary complexity of gates in liquid crystal and chemical medium is not possible at present.

We have based our conclusions only on two experimental studies, conducted by independent groups of researcher. The studies are far from being complete, and results are exposed to interpretations. Thus, when talking about liquid crystal we constructed hierarchy of evolutionary complexity based on minimal number of evaluations [Harding and Miller, 2005]. If we would consider averages then most behaviourally complex gate xor would be the easiest to evolve:

NOT ◂ XOR ◂ OR ◂ NAND ◂ NOR ◂ AND

When dealing with evolution of gates in Belousov-Zhabotinsky system we selected results based on mutation highest mutation rate 6000 [Toth et al., 2007]. If we would choose lower mutation rate 4000 then we would finish with evolutionary hierarchy:

AND ◂ XOR ◂ NAND

We aim to achieve the consistency in our future works by analysing more experimental data.

With regard to methodology we have embraced the following approach. A non-linear physico-chemical system is evolved to implement a logical function. Then the logical function is considered to be a cell state transition of some cellular automata. Complexity of the cellular automata was compared to complexity of evolution. The next step could be to evaluate complexity of the non-linear media themselves. Is liquid crystal less complex than Belousov-Zhabotinsky medium? This may be a subject for future studies.

**FIGURES**

| Gate | Min. Evals. | Max. Evals. | Avg. Eval.s | Std. dev |
|---|---|---|---|---|
| AND | 2 | 1788 | 910 | 527.45 |
| OR | 1 | 1779 | 769 | 576.01 |
| XOR | 44 | 1255 | 649 | 605.50 |
| NOT | 3 | 1750 | 536 | 749.58 |
| NAND | 13 | 1763 | 880 | 623.18 |
| NOR | 1 | 1788 | 907 | 526.97 |

(a)

| Gate | Controller | Mutation rate | Success rate | Min. gens | Max. gens | Avg. gens | Std.dev. |
|---|---|---|---|---|---|---|---|
| | Coevolutionary | 4000 | 10/10 | 8 | 144 | 61 | 45.69 |
| | Random | 4000 | 10/10 | 4 | 200 | 64 | 71.73 |
| | Coevolutionary | 6000 | 10/10 | 16 | 84 | 49 | 21.89 |
| AND | Random | 6000 | 10/10 | 8 | 176 | 66 | 58.64 |
| | Coevolutionary | 4000 | 7/10 | 288 | >2000 | 1065 | 767.21 |
| | Random | 4000 | 4/10 | 300 | >2000 | 1454 | 744.49 |
| | Coevolutionary | 6000 | 9/10 | 24 | >2000 | 847 | 829.22 |
| NAND | Random | 6000 | 6/10 | 84 | >2000 | 1247 | 727.81 |
| | Coevolutionary | 4000 | 9/10 | 348 | >2000 | 808 | 510.08 |
| | Random | 4000 | 10/10 | 20 | 1080 | 455 | 333.68 |
| | Coevolutionary | 6000 | 9/10 | 32 | >2000 | 1118 | 574.97 |
| XOR | Random | 6000 | 6/10 | 212 | >2000 | 1336 | 635.84 |

(b)

Fig. 1. Evolutionary complexity of logic gates. (a) Characteristics of evolving logic gates in liquid crystal [Harding & Miller, 2005]. (b) Characteristics of evolving logic gate in light-sensitive Belousov-Zhabotinsky medium [Toth et al., 2007]

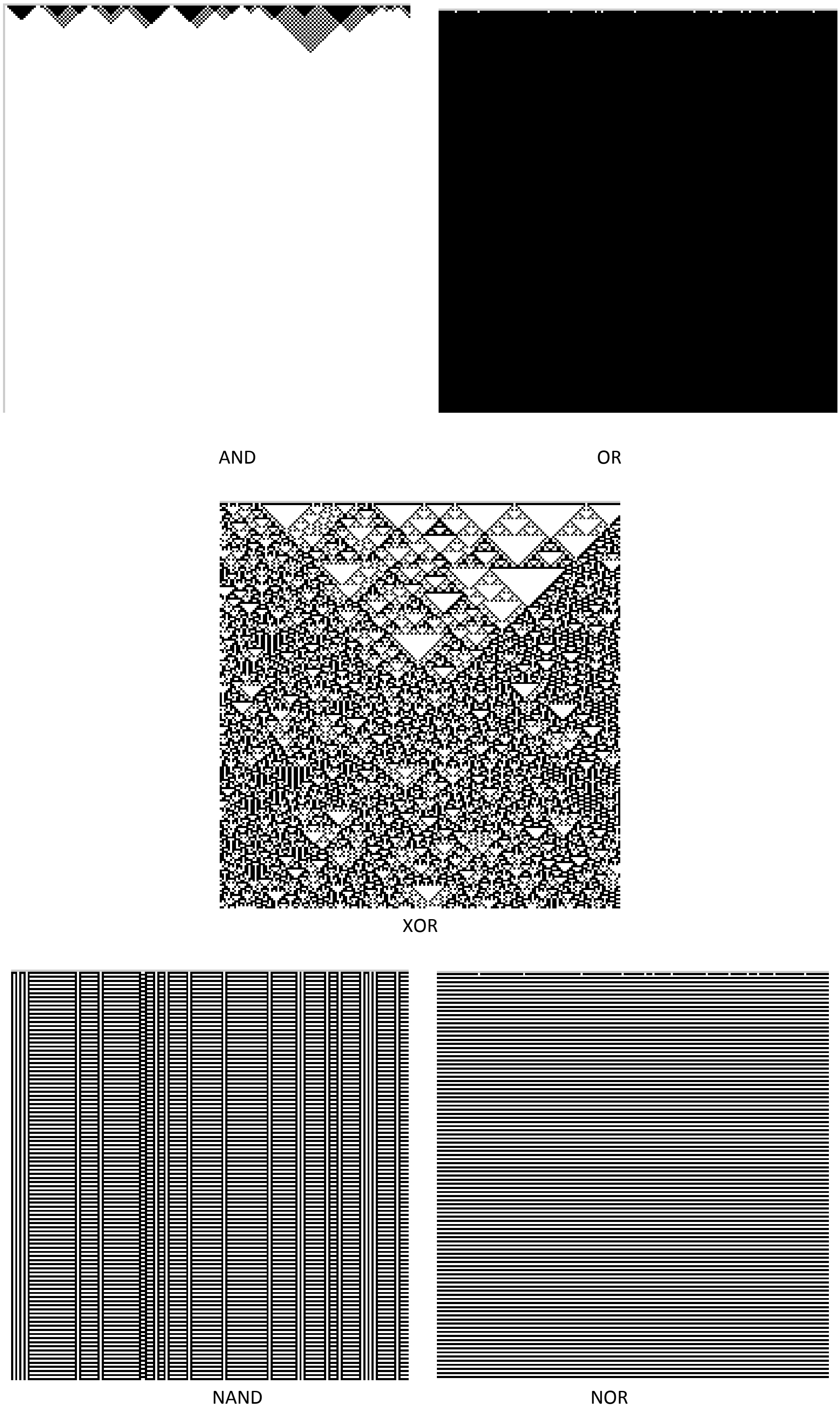

**Fig. 2.** Space-time configurations (200 cells by 200 time steps) of one-dimensional cellular automata, where each cell plays a role of a logic gate with input from ‘left’ and ‘right’ neighbours. Each automaton has 200 cells. Time goes down. Cells in state TRUTH are shown by black pixels, FALSE by blanks.

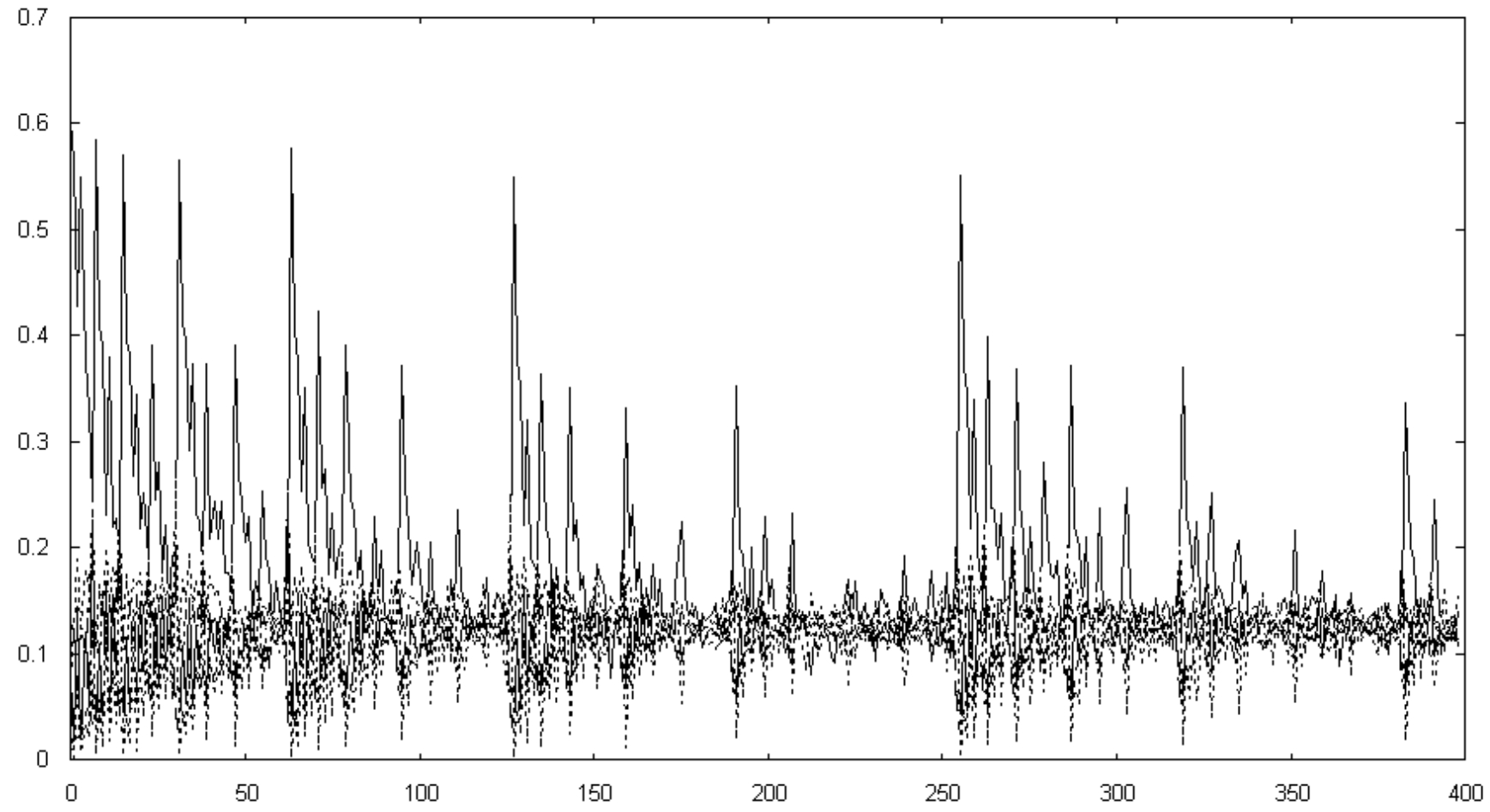

(a)

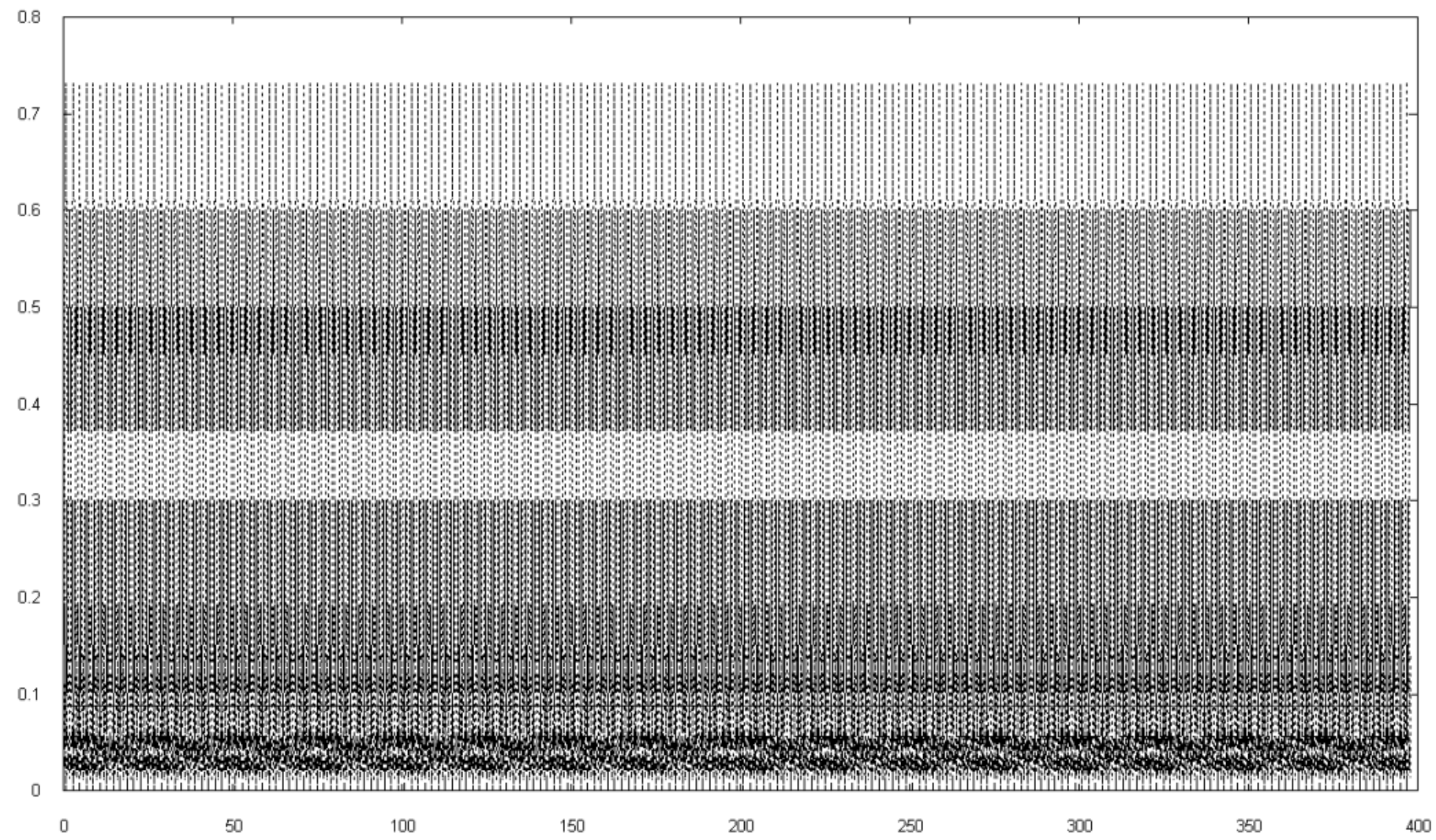

(b)

**Fig. 3.** Frequencies of neighbourhood states in configurations of cellular automata governed by (a) XOR and (b) NAND cell-state transition rules.